\documentclass[]{ceurart}
\usepackage{graphics}
\usepackage{xcolor}

\begin{document}

\copyrightyear{2021}
\copyrightclause{Copyright for this paper by its authors.
  Use permitted under Creative Commons License Attribution 4.0
  International (CC BY 4.0).}

\conference{CLEF 2021 -- Conference and Labs of the Evaluation Forum, 
	September 21--24, 2021, Bucharest, Romania}

\title{Early Risk Detection of Pathological Gambling, Self-Harm and Depression Using BERT}

\author[1]{Ana-Maria Bucur}[
orcid=0000-0003-2433-8877,
email=ana-maria.bucur@drd.unibuc.ro,
]
\address[1]{Interdisciplinary School of Doctoral Studies, University of Bucharest}

\author[2]{Adrian Cosma}[%
orcid=0000-0003-0307-2520,
email=cosma.i.adrian@gmail.com,
]
\address[2]{Politehnica University Bucharest}

\author[3, 4]{Liviu P. Dinu}[%
orcid=0000-0002-7559-6756,
email=liviu.p.dinu@gmail.com,
]
\address[3]{Faculty of Mathematics and Computer Science, University of Bucharest}
\address[4]{Human Language Technologies Research Center, University of Bucharest}

\begin{abstract}
    Early risk detection of mental illnesses has a massive positive impact upon the well-being of people. The eRisk workshop has been at the forefront of enabling interdisciplinary research in developing computational methods to automatically estimate early risk factors for mental issues such as depression, self-harm, anorexia and pathological gambling. In this paper, we present the contributions of the BLUE team in the 2021 edition of the workshop, in which we tackle the problems of early detection of gambling addiction, self-harm and estimating depression severity from social media posts. We employ pre-trained BERT transformers and data crawled automatically from mental health subreddits and obtain reasonable results on all three tasks. 
\end{abstract}

\begin{keywords}
    deep learning\sep
    mental health \sep
    early risk detection \sep
    social media \sep
    transfer learning \sep
    weakly-supervised learning \sep
    natural language processing
\end{keywords}

\maketitle

\section{Introduction}
Mental disorders are a prevalent problem of our world, 10.7\% of people worldwide live with a mental health disorder\footnote{https://ourworldindata.org/mental-health}. There are various mental illnesses such as depression, anxiety, bipolar disorder, eating disorders or addictions (including alcohol, drugs or gambling). For many people living with these problems, it is hard to access treatment because of the social stigma, many mental illnesses remain undiagnosed.

With the rise in social media popularity, more and more researchers are using it as a source for gathering data because users tend to talk about their emotions and mental health problems online and to seek support. 

According to the DSM-5, pathological gambling is a mental health illness in which a person has a persistent and repetitive problematic gambling behaviour. They are constantly preoccupied with gambling, feel a need to spend more money, want to control or to quit their gambling problem, but already have several unsuccessful attempts, they try to hide their addiction and rely on others to regain a stable financial status or suffer other consequences \cite{american2013diagnostic}. For pathological gambling screening, several instruments have been developed, such as The Lie/Bet Questionnaire \cite{johnson1997lie}, The Gambling Urge Scale \cite{raylu2004gambling} and The Gambling Craving Scale \cite{young2009gambling}. 
Besides the classical gambling methods, with the rise of the Internet, the availability of gambling opportunities has increased. With the high accessibility of Internet gambling, the incidence and prevalence of pathological gambling has also increased \cite{gainsbury2011internet, gainsbury2015online}. 

Self-harm is a serious mental health problem, people suffering from this condition often also being diagnosed for other mental health disorders: borderline personality disorder, depression, anxiety, eating disorders, post-traumatic stress disorder (PTSD), addictions (alcohol and drugs) \cite{fliege2009risk}. Deliberate self-harm behaviour refers to the act of self-inducing harming and resulting in tissue damage. It does not include other types of damage to the body (not exercising, unhealthy eating patterns, engaging deliberately in actions resulting in psychological self-harm). Regarding adolescents, 17\% had been involved in non-suicidal self-injury at least once. However, self-harm is less common in adults, with only 5\% of them engaging in this behaviour\footnote{\url{https://www.apa.org/monitor/2015/07-08/who-self-injures}}.

Depression is the leading cause of disability, with more than 264 million people from all age groups suffering from this disorder and women being more affected than men\footnote{https://www.who.int/news-room/fact-sheets/detail/depression}. For measuring the severity of depression, one of the most widely used instruments is the Beck Depression Inventory (BDI) \cite{Upton2013}, a 21-question multiple-choice questionnaire that can assess depression symptoms from the patient's own experience. This questionnaire is a powerful tool that allows for a rapid diagnosis of one's mental state. Moreover, through computational methods, the answers to the questions can be estimated by looking at their social media interactions and writings.

The early risk detection task became a popular topic in the interdisciplinary research area combining natural language processing and psychology. The Early Risk Prediction on the Internet (eRisk) Workshop\footnote{https://erisk.irlab.org/}, part of the Conference and Labs of the Evaluation Forum (CLEF), focuses on the effectiveness and practical applications of early risk detection of different pathologies or safety threats. In addition, the workshop explores the technologies that can be used for people's health and safety, and the strategies related to building test collections \cite{losada2020overview}. 

Throughout the five editions of the eRisk workshop, the tasks addressed pathological gambling, depression, self-harm and anorexia. Another workshop tackling the early detection task is the Workshop on Computational Linguistics and Clinical Psychology (CLPsych)\footnote{https://clpsych.org/} with tasks such as predicting current and future psychological health from childhood essays \cite{lynn2018clpsych} and early detection of suicide risk \cite{zirikly2019clpsych}. Other efforts on early risk detection are also being made for the early detection of cognitive decline \cite{beltrami2018speech} or antisocial behaviour \cite{singh2019framework}.

In this edition of the eRisk workshop, 17 teams developed their models to solve the proposed tasks, the highest number of participating teams in the workshop's history. \cite{parapar2021overview}
The tasks from the current edition of the workshop are:
\begin{itemize}
    \item Task 1: Early Detection of Signs of Pathological Gambling
    \item Task 2: Early Detection of Signs of Self-Harm
    \item Task 3: Measuring the Severity of the Signs of Depression
\end{itemize}

With the recent advancements in deep learning and natural language processing, we focus our solutions by making use of state-of-the-art pre-trained transformer models \cite{devlin-etal-2019-bert} (i.e. BERT) and scraping publicly available data from social media websites. Further, we make an overview of existing methods from previous editions of the eRisk workshop and present our solutions for Tasks 1, 2 and 3.

\section{Related Work}
Automated methods for compulsive gambling detection focus on analyzing behavioural player data (frequency of betting, number of different games played, etc.) \cite{adami2013markers, peres2021time}, transactional data \cite{ladouceur2017responsible} or texts from the correspondence between customers and customer service employees from online gambling operators \cite{haefeli2011early}. To the best of our knowledge, the eRisk workshop is the first to tackle pathological gambling detection from social media text.

The task on early detection of signs of self-harm is a continuation of eRisk 2019's T2 and 2020's T1 tasks. The best performing runs from the previous workshop edition in terms of Precision, $F$1 and ${ERDE}$ were developed by \citet{martinez2020early}. The team used for sentence classification models based on XLM-RoBERTa \cite{conneau2019unsupervised} trained on different datasets. They avoided using the training dataset given by the eRisk organizers and instead gathered their own dataset from the Pushshift Reddit Dataset \cite{baumgartner2020pushshift}. For the positive group, all posts from users who posted or commented in \textit{r/selfharm} subreddit were extracted, including their submissions to other subreddits. A small proportion of user posts were manually annotated by the team, and the rest of them were used as positive class without further labeling. For the control group, they extracted random users who have never posted on a self-harm-related subreddit. Following this approach, the team obtained a small dataset, manually labeled, and three other larger datasets automatically generated. In their experiments, they used only the last three collections. Other approaches to self-harm detection were using bag-of-words \cite{oliveira2020bioinfo, achilles2020using}, bag of sub-emotions \cite{aragon2020inaoe}, LIWC, emotion and sentiment features \cite{uban2020deep} or deep learning models \cite{losada2020overview}. The results from this task suggest that the systems can obtain good performance on very little data and can be used to support human experts in the early detection of the signs of self-harm  \cite{losada2020overview}.
 
The task of measuring the severity of the signs of depression is a continuation of eRisk 2019's T3 and 2020's T2 tasks. The best performing run from the last edition in terms of the Average Hit Rate (AHR) metric was the one of \cite{oliveira2020bioinfo}. The team used a machine learning model for depression detection from their previous participation in the eRisk workshop, trained on the RSDD dataset \cite{DBLP:journals/corr/abs-1709-01848}. They predicted if a user was depressed or not with this model and they conjugated the score obtained with several psycholinguistics and behavioural features, such as polarity, use of self-related words, use of absolutist words, mentions of mental disorders-related terms and other terms regarding the symptoms of depression from the BDI questionnaire. Regarding the psycholinguistics features, \citet{bucur2021exploratory} showed through cross-dataset analysis the connection between offensive language and signs of depression. Other approaches were based on transformers models \cite{martinez2020early}, SVM and logistic regression \cite{uban2020deep}, LDA \cite{maupome2020early}, writing style \cite{maupome2020early}, etc. 

The results from this task show that a depressive screening tool that can predict the answers to the BDI questionnaire from social media posts has not reached a good performance that allows it to be used in a real scenario \cite{losada2020overview}, as opposed to the previous task on early detection on signs of depression which had good results when evaluating the error measure for early risk detection \cite{losada2018overview, funez2018unsl, bucur2020detecting}

\section{Task 1: Early Detection of Signs of Pathological Gambling}
The first task of this edition of the eRisk workshop is detecting early signs of pathological gambling. This is a novel task, not appearing in other workshop editions, which focuses on data mining from social media. As such, the organizers do not provide a training set and only offer a testing platform. The data from this task is collected from Reddit, following the same methodology described by \citet{losada2016test}. All the data used in this workshop was gathered from Reddit, as the social media platform allows the use of its' public data for research purposes \cite{losada-crestani2016}. The data is a collection of users posts from different subreddits.

In 2019, the workshop switched from a chunk-based release of the data to a sequence-based approach for the test stage. Before 2019, the participants tested their runs on 10 chunks with several submissions from each user, which were released every week. In the current workshop, each team had to process their runs only on one post at a time from each user in chronological order. 

Our approach is to construct a large dataset of user posts, loosely annotated, such that we can train our model. We follow a methodology similar to \citet{martinez2020early} and use the posts from the \textit{r/GamblingAddiction}\footnote{\url{https://www.reddit.com/r/GamblingAddiction/}} and \textit{r/problemgambling}\footnote{\url{https://www.reddit.com/r/problemgambling/}} subreddits as positive training data for pathological gambling detection. Posts in these subreddits are mostly discussing gambling addiction. We assume that most of the regular users in this subreddit are pathological gamblers. While it is evident that not all users from this subreddit have a gambling addiction, it is sufficient to train our model.

As such, we crawled the gambling-related subreddits and obtained 2111 unique users with a total of 13377 posts only in these two subreddits. To avoid having only posts containing topics related to gambling, from the 2111 users we also crawled their posts from other unrelated subreddits, which increased the posts from the positive class to 80998.

For the collection of control users, we make use of the control split of the Reddit Self-reported Depression Diagnosis (RSDD) \cite{DBLP:journals/corr/abs-1709-01848} and Early Risk Prediction on the Internet (eRisk) 2018 \cite{losada-crestani2016} depression datasets. RSDD and eRisk
are two publicly available datasets that contain posts also gathered from the Reddit social media platform. Both datasets contain users labeled as having depression by their mention of diagnosis in their posts and control users. In RSDD, each user in the depression group was matched with 12 control users who has the most similar post probability distributions. The control group contains users who do not suffer from depression, there is not any mention of diagnosis in their posts, they do not have depression-related terms in their posts and they had never posted in a mental health-related subreddit. For eRisk, the control group is composed of random Reddit users, some of them active in mental health-related subreddits, but without any mention of diagnosis.

The control splits used in our experiments contain random posts from users from various subreddits and the number of posts was chosen to match the number of posts in the positive class. We chose two variants, one that uses the control split from RSDD and one that uses the control split from eRisk 2018. We wanted to see the impact of the two different control datasets, as RSDD does not contain posts related to mental health, but eRisk does include posts from users active in mental health subreddits.

Further, we fine-tuned a pre-trained BERT classifier using AdamW optimizer with a learning rate of 0.0002 for 2 epochs. For prediction, we chose aggressive thresholds of 0.99 / 0.98 on the output probability to decide whether or not the user presents signs of gambling addiction. This high threshold minimizes false positives and ensures that a decision is taken only if a post denotes signs of addiction.


For this task, evaluation is performed similarly to previous eRisk workshops, namely Precision, Recall, F1,  the ERDE \cite{Trotzek_2020} metric as an early risk detection measure and F$_{latency}$ \cite{10.1145/3159652.3159725}. F$_{latency}$ was deemed more interpretable and informative for this task as it combines the effectiveness of the decision (estimated with the F measure) and the delay. Testing data for Task 1 was released on an item-by-item basis through an API that processed current predictions for all runs, and progressively released the next sequence data. We opted for 5 runs and processed 1828 user writings, for a total of 1 day and 23:43:28 hours.

The five runs and the corresponding datasets used for the classification of pathological gamblers are described below.

\textbf{RUN 0} BERT model trained on all the posts (including those from unrelated subreddits) from users who posted in the gambling-related subreddits, with control data from eRisk, with a 0.99 threshold

\textbf{RUN 1} BERT model trained on all the posts (including those from unrelated subreddits) from users who posted in the gambling-related subreddits, with control data from RSDD, with a 0.99 threshold

\textbf{RUN 2} BERT model trained on all the posts from the gambling subreddits with control data from eRisk, with a 0.99 threshold for classification

\textbf{RUN 3} BERT model trained on all the posts (including those from unrelated subreddits) from users who posted in the gambling-related subreddits, with control data from eRisk, with a 0.98 threshold

\textbf{RUN 4} BERT model trained on all the posts from the gambling subreddits with control data from eRisk, with a 0.98 threshold for classification

\begin{table}[hbt!]
    \centering
    \resizebox{\textwidth}{!}{
        \begin{tabular}{lccccccccc}
        \toprule
        team name & run id & P & R & F1 & ERDE$_5$ & ERDE$_{50}$ & latency$_{TP}$ & speed & latency-weighted F1 \\
        \toprule
        BLUE & 0 & .107 & .994 & .193 & .067 & .046 & 2 &  .996 & .192 \\
        BLUE & 1 & .157 & .988 & .271 & .054 & .036 & 2 &  .996 & .270 \\
        BLUE & 2 & .121 & .994 & .215 & .065 & .045 & 2 &  .996 & .215 \\
        BLUE & 3 & .095 & \textbf{1} & .174 & .071 & .051 & 2 &  .996 & .173 \\
        BLUE & 4 & .110 & .994 & .198 & .068 & .048 & 2 &  .996 & .197 \\
        \midrule
        UNSL & 2 & \textbf{.586} & .939 & \textbf{.721} & .073 & \textbf{.020} & 11 & .961 & \textbf{.693} \\
        RELAI & 0 & .138 & .988 & .243 & \textbf{.048} & .036 & 1 &  \textbf{1} & .243 \\
        \bottomrule
        \end{tabular}
    }
    \caption{Decision-based evaluation on Task 1: Early Detection of Signs of Pathological Gambling. }
    \label{tab:task1-results}
\end{table}

\begin{table}[hbt!]
    \centering
    \resizebox{\textwidth}{!}{
        \begin{tabular}{lccccccccccccccc}
        \toprule
        & & \multicolumn{3}{c}{1 writing} & \multicolumn{3}{c}{100 writings} & \multicolumn{3}{c}{500 writings} & \multicolumn{3}{c}{1000 writings} \\
        team & run & P@10 & NDCG@10 & NDCG@100 & P@10 & NDCG@10 & NDCG@100 & P@10 & NDCG@10 & NDCG@100 & P@10 & NDCG@10 & NDCG@100 \\
        \toprule
        BLUE & 0 & .9 & .88 & .61 & .8 & .73 & .57 & .9 & .93 & .64 & .7 & .78 & .60 \\
        BLUE & 1 & \textbf{1} & \textbf{1} & .61 & .8 & .82 & .53 &  \textbf{1} & \textbf{1} & .56 & \textbf{1} & \textbf{1} & .56 \\
        BLUE & 2 & .6 & .70 & .73 & .8 & .87 & .76 & .8 & .88 & .75 & .9 & .90 & .76 \\
        BLUE & 3 & .6 & .65 & .60 & .8 & .87 & .61 & .7 & .71 & .60 & .7 & .67 & .60 \\
        BLUE & 4 & .9 & .81 & .73 & \textbf{1} & \textbf{1} & .77 & \textbf{1} & \textbf{1} & .76 & \textbf{1} & \textbf{1} & .78 \\
        \midrule
        UNSL & 2 & \textbf{1} & \textbf{1} & \textbf{.85} & \textbf{1} & \textbf{1} & \textbf{1} & \textbf{1} & \textbf{1} & \textbf{1} & \textbf{1} & \textbf{1} & \textbf{1} \\
        \bottomrule
        \end{tabular}
    }
    \caption{Ranking-based evaluation on Task 1: Early Detection of Signs of Pathological Gambling. }
    \label{tab:task1-results-ranking}
\end{table}

Our results, compared with the best results from the early detection of pathological gambling task, are presented in Table \ref{tab:task1-results} for the decision-based evaluation and Table \ref{tab:task1-results-ranking} for the ranking-based evaluation. Our runs obtained high Recall, but high Precision and F1 are also needed in a real-life scenario on problem gambling detection. Our ERDE scores were relatively close to the best result from this task. Regarding the ranking-based evaluation, our team's ratings got better as our models analyzed more data from each user. Even after one writing from each user, the model was quite effective at predicting the risk of users and prioritizing them.

\section{Task 2: Early Detection of Signs of Self-Harm}
For the second task, on early self-harm detection, the organizers provided a training dataset composed of all 2020's T1 users (both training and test) plus the test users of 2019's T2 task. This dataset contains 144 users in the self-harm class and 608 users in the control class, with a total of 120,898 and 1,241,271 posts, respectively. However, we increased the size of this dataset by crawling the \textit{r/selfharm} subreddit. This subreddit contains various posts from users seeking help and stories related to their own experiences. We deemed the users from this subreddit as being at risk of self-harm. We chose two variants for control users, one that uses the control split from RSDD and one that contains the control split from eRisk 2018, as in the previous task on pathological gambling detection.

We followed the same methodology as in the previous task. We also use a BERT model trained with the same parameters. For prediction, we chose aggressive thresholds of 0.99 / 0.98 / 0.95 on the output probability to decide whether or not the user presents signs of self-harm. This high threshold minimizes false positives and ensures that a decision is taken only if a post denotes signs of self-harm.


Similarly to Task 1, testing data was released progressively after all runs offered predictions for the current batch of posts. We opted for 5 runs, but processed only 156 user writings in 1 day and 4:57:23 hours, due to the lack of computational resources and server latency.

\begin{table}[hbt!]
    \centering
    \resizebox{\textwidth}{!}{
        \begin{tabular}{lccccccccc}
        \toprule
        team name & run id & P & R & F1 & ERDE$_5$ & ERDE$_{50}$ & latency$_{TP}$ & speed & latency-weighted F1 \\
        \toprule
        BLUE & 0 & .283 & .934 & .435 & .084 & .041 & 5 & .984 & .428  \\
        BLUE & 1 & .142 & .875 & .245 & .117 & .081 & 4 & .988 & .242  \\
        BLUE & 2 & .454 & .849 & .592 & .079 & .037 & 7 & .977 & .578  \\
        BLUE & 3 & .394 & .868 & .542 & .075 & .035 & 5 & .984 & .534  \\
        BLUE & 4 & .249 & .928 & .393 & .085 & .044 & 4 & .988 & .388  \\
        \midrule
        Birmingham & 2 & \textbf{.757} & .349 & .477 & .085 & .07 & 4 & .988 & .472  \\
        UPV-Symanto & 1 & .276 & .638 & .385 & \textbf{.059} & .056 & 1 & 1 &.0 .385 \\
        UNSL & 0 & .336 & .914 & .491 & .125 & \textbf{.034} & 11 & .961 & .472  \\
        UNSL & 4 & .532 & .763 & \textbf{.627} & .064 & .038 & 3 & .992 & \textbf{.622} \\
        \bottomrule
        \end{tabular}
    }
    \caption{Decision-based results for Task 2: Early Detection of Signs of Self-Harm. }
    \label{tab:task2-results}
\end{table}

\begin{table}[hbt!]
    \centering
    \resizebox{\textwidth}{!}{
        \begin{tabular}{lccccccccccccccc}
        \toprule
        & & \multicolumn{3}{c}{1 writing} & \multicolumn{3}{c}{100 writings} & \multicolumn{3}{c}{500 writings} & \multicolumn{3}{c}{1000 writings} \\
        team & run & P@10 & NDCG@10 & NDCG@100 & P@10 & NDCG@10 & NDCG@100 & P@10 & NDCG@10 & NDCG@100 & P@10 & NDCG@10 & NDCG@100 \\
        \toprule
        BLUE & 0 & 0.7 & 0.75 & 0.54 & 0.8 & 0.82 & 0.59 & 0.0 & 0.0 & 0.0 & 0.0 & 0.0 & 0.0 \\
        BLUE & 1 & 0.2 & 0.13 & 0.26 & 0.4 & 0.41 & 0.29 & 0.0 & 0.0 & 0.0 & 0.0 & 0.0 & 0.0 \\
        BLUE & 2 & 0.6 & 0.49 & 0.5 & \textbf{0.9} & \textbf{0.94} & 0.55 & 0.0 & 0.0 & 0.0 & 0.0 & 0.0 & 0.0 \\
        BLUE & 3 & 0.6 & 0.43 & 0.49 & 0.8 & 0.87 & 0.54 & 0.0 & 0.0 & 0.0 & 0.0 & 0.0 & 0.0 \\
        BLUE & 4 & 0.7 & 0.61 & 0.52 & 0.8 & 0.88 & 0.55 & 0.0 & 0.0 & 0.0 & 0.0 & 0.0 & 0.0 \\
        \midrule
        UPV-Symanto & 3 & 0.6 & 0.7 & 0.51 & \textbf{0.9} & \textbf{0.94} & 0.69 & 0.9 & 0.94 & 0.69 & 0.0 & 0.0 & 0.0 \\
        UNSL & 0 & \textbf{1} & \textbf{1} & \textbf{0.7} & 0.7 & 0.74 & \textbf{0.82} & 0.8 & 0.81 & 0.8 & 0.8 & 0.81 & \textbf{0.8} \\
        UNSL & 3 & \textbf{1} & \textbf{1} & 0.63 & \textbf{0.9} & 0.81 & 0.76 & 0.9 & 0.81 & \textbf{0.71} & 0.8 & 0.73 & 0.69 \\
        \bottomrule
        \end{tabular}
    }
    \caption{Ranking-based evaluation on Task 1: Early Detection of Signs of Pathological Gambling. }
    \label{tab:task2-results-ranking}
\end{table}

The five runs and the corresponding datasets used for the classification of self-harmers are described below.

\textbf{RUN 0} BERT model trained only on the training data provided by the organizers, with a 0.95 threshold

\textbf{RUN 1} BERT model trained on all the posts (including those from unrelated subreddits) from users who posted in the self-harm subreddit, with control data from eRisk, with a 0.99 threshold

\textbf{RUN 2} BERT model trained on the training data provided by the organizers and posts from the self-harm subreddit, with control data from eRisk, with a 0.99 threshold

\textbf{RUN 3} BERT model trained on the training data provided by the organizers and posts from the self-harm subreddit, with control data from eRisk, with a 0.98 threshold

\textbf{RUN 4} BERT model trained on the training data provided by the organizers and posts from the self-harm subreddit, with control data from eRisk, with a 0.95 threshold

Our results, and the best results, for the early detection task of pathological gambling are presented in Table \ref{tab:task2-results} for the decision-based evaluation and Table \ref{tab:task2-results-ranking} for the ranking-based evaluation. Our team obtained good results in terms of F1 and ERDE measures. For the ranking-based evaluation, we obtained the best P@10 and NDCG@10 scores after 100 writings, among other teams, which means that these runs were effective to detect the risk of self-harm of users.

\section{Task 3: Measuring the severity of the signs of depression}
For this task, participants are required to estimate the level of depression of a user, given their history of postings. Users have filled in the Beck's Depression Inventory (BDI) \cite{Upton2013}, which assesses the presence of feelings of sadness, pessimism, loss of energy. The questionnaire is comprised of 21 questions, with 4 answers each, denoting the severity of their respective symptom of depression, except for questions 16 and 18, which have 6 answers each (0, 1a, 1b, 2a, 2b, 3a, 3b). Questions 16 and 18 refer to changes in sleeping patterns and appetite, in which both extremes (undersleeping or oversleeping, for instance) tell of one's mental health condition.

For model training, data from the 2019 and 2020 challenges was provided, which contains a total of 89 users with ground truth questionnaire answers and a total of 46502 posts from Reddit.

We are provided with 80 users and their posts for evaluation, totalling 32237 posts and comments from Reddit. The evaluation for this task is threefold: number of correct responses, the absolute difference between levels of depression obtained from the ground truth responses to the questionnaire vs the estimated answers, and the depression level, which is used to categorize users as \textit{minimal depression} (0-9), \textit{mild depression} (10-18), \textit{moderate depression} (19-29), \textit{severe depression} (30-63).

The number of correct responses is measured by the \textbf{Average Hit Rate} (AHR). \textbf{Average Closeness Rage} (ACR) takes into account the responses of the depression questionnaire represent an ordinal scale, not only separate options. \textbf{Average Difference between Overall Depression Levels} (ADODL) measure computes the overall depression level for the ground truth and automated questionnaire. Finally, \textbf{Depression Category Hit Rate} (DCHR) calculates the fraction of cases where the automated questionnaire led to a depression category that is equivalent to the depression category obtained from the ground truth questionnaire \cite{parapar2021overview}.

\begin{figure}
    \centering
    \includegraphics[width=0.8\textwidth]{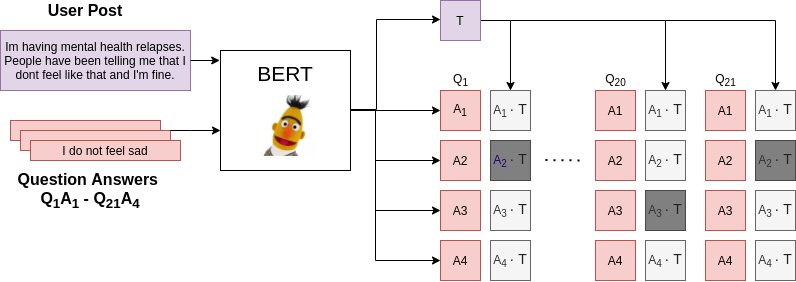}
    \caption{Our architecture for determining answers to the BDI questionnaire. We use a BERT encoder that encodes both the user text and all the answers to the questionnaire. We then fine-tune the BERT encoder through contrastive learning. \textit{Note:} Exemplified user post is paraphrased for privacy reasons.}
    \label{fig:task-3-arch}
\end{figure}

To determine the answers to the questionnaire, we trained several BERT models. Firstly, we fine-tuned a BERT classifier having 21 x 4 outputs, corresponding to all the 84 possible responses to the questionnaire. Each question \textit{head} outputs a probability distribution across its possible answers. For questions 16 and 18, we merged answers. The model was trained using cross-entropy loss, with a learning rate of 0.0002 for 2 epochs, using AdamW optimizer.

For our second run, we added another classification \textit{head}, which outputs a probability distribution over the severity score (minimal/mild/moderate/severe). The training regime was identical to our first scenario.

Finally, we constructed an architecture that takes into account each answer's natural language formulation. At each training iteration, a user post and all the answers are fed into the BERT model, and the output is given by the dot product between embedding vectors of the post text and the answer text. Both answer embeddings the user text embedding are trained alongside. Figure \ref{fig:task-3-arch} shows the diagram of our architecture. 
This approach is similar to recent proposals to leverage contrastive learning as a way to train neural networks, especially in noisy or scarce data regimes. By computing the similarity between the answers and the user posts, we make use of the semantic meaning of the answer text, which provides more informed predictions than plain classification.  The network was trained using AdamW optimizer for 2 epochs with a learning rate of 0.00002.

Since the dataset is noisy, and not all posts contain relevant information for all questions, at test time, we select the answer with the maximum probability across all user posts for a particular user. 

\begin{table}[hbt!]
    \centering
    \resizebox{\textwidth}{!}{
        \begin{tabular}{lcccc}
        \toprule Run & AHR & ACR & ADODL & DCHR \\
        \toprule BLUE run0 & $27.86 \%$ & $64.66 \%$ & $74.15 \%$ & $17.50 \%$ \\
            BLUE run1 & $30.00 \%$ & $64.58 \%$ & $70.65 \%$ & $11.25 \%$ \\
            BLUE run2 & $30.36 \%$ & $65.42 \%$ & $75.42 \%$ & $21.25 \%$ \\
            BLUE run3 & $29.52 \%$ & $64.70 \%$ & $73.63 \%$ & $13.75 \%$ \\
            BLUE run4 & $29.76 \%$ & $65.04 \%$ & $74.84 \%$ & $15.00 \%$  \\
        \midrule
            CYUT run2 & 32.62\% & 69.46\% & \textbf{83.59}\% & \textbf{41.25}\% \\
            DUTH\_ATHENA MeanPostsSVM & \textbf{35.36}\% & 67.18\% & 73.97\% & 15.00\% \\
            UPV-Symanto 4\_symanto\_upv\_lingfeat\_cors & 34.17\% & \textbf{73.17}\% & 82.42\% & 32.50\% \\
        \bottomrule
        \end{tabular}
    }
    \caption{Competition results for \textbf{Task 3}: Measuring the severity of the signs of depression. Best results are obtained through similarity learning.}
    \label{tab:task3-results}
\end{table}

Table \ref{tab:task3-results} showcases our results on Task 3, compared to the best results from other teams. The five runs are described below.

\textbf{RUN 0} corresponds to the plain classification of answers

\textbf{RUN 1} is the joint prediction of responses and depression severity

\textbf{RUN 2} is the similarity learning architecture

\textbf{RUN 3} is an ensemble of the outputs from runs 0, 1 and 2

\textbf{RUN 4} is an ensemble of the outputs from runs 0 and 2

Our best run (run 2) has the best overall results, showcasing that leveraging the question context is an important step towards accurate depression screening. Better results will undoubtedly be obtained if one considers only posts that are relevant to a particular question.

\section{Conclusion}
In this paper, we presented the contributions of the BLUE team in the eRisk 2021 workshop on early risk detection on the internet. Our approaches tackled early risk detection of pathological gambling, self-harm and estimating depression severity from a user's social media posts (i.e. Reddit). For early risk detection of problem gambling, we opted for crawling and constructing a dataset from popular subreddits that address gambling addiction and trained a BERT classifier on user posts, with control users from RSDD and eRisk 2018, two popular datasets for depression detection. Similarly, we crawled the subreddit associated with the self-harm behaviour to augment the existing training dataset provided by the competition organizers for the detection of self-harm. Finally, for estimating depression severity, we opted for constructing an architecture that combines the questionnaire answers with a user's posts to estimate the probability of correct answers. We obtained reasonable results through our methods. However, these problems are still far from being solved, and further research is needed to make computational methods feasible for use in the real world. The positive impact of early risk detection for various mental health disorders cannot be emphasized enough, and has the potential to save many human lives.

\bibliography{refs}

\end{document}